\documentclass[a4paper, 10pt, conference]{ieeeconf}      

\IEEEoverridecommandlockouts                              

\usepackage[dvipdfmx]{graphicx}
\usepackage{comment}

\title{\LARGE \bf
Interactively Robot Action Planning with Uncertainty Analysis \\ and Active Questioning by Large Language Model
}

\author{Kazuki Hori$^{1}$, Kanata Suzuki$^{1,2}$, Tetsuya Ogata$^{1,3,4}$
\thanks{
$^{1}$Kazuki Hori, $^{1,2}$Kanata Suzuki, and $^{1,3,4}$Tetsuya Ogata are with Faculty of Science and Engineering, Waseda University, Tokyo 169-8050, Japan. 
$^{1,2}$Kanata Suzuki is also with Artificial Intelligence Laboratories, Fujitsu Limited., Kanagawa 211-8588, Japan. 
$^{1,3,4}$Tetsuya Ogata is also with the Waseda Research Institute for Science and
Engineering (WISE) at Waseda University, Tokyo 169-8555, Japan, and the National Institute of Advanced Industrial Science and Technology, Tokyo 100-8921, Japan.
E-mail: {\tt\small ogata@waseda.jp}
.}
}

\begin{document}

\maketitle
\thispagestyle{empty}
\pagestyle{empty}

\begin{abstract}
The application of the Large Language Model (LLM) to robot action planning has been actively studied. The instructions given to the LLM by natural language may include ambiguity and lack of information depending on the task context. It is possible to adjust the output of LLM by making the instruction input more detailed; however, the design cost is high. In this paper, we propose the interactive robot action planning method that allows the LLM to analyze and gather missing information by asking questions to humans. The method can minimize the design cost of generating precise robot instructions. We demonstrated the effectiveness of our method through concrete examples in cooking tasks. However, our experiments also revealed challenges in robot action planning with LLM, such as asking unimportant questions and assuming crucial information without asking. Shedding light on these issues provides valuable insights for future research on utilizing LLM for robotics.
\end{abstract}


\section{Introduction}
In recent years, the development of Large Language Models (LLMs~\cite{Zhao2023}) has led to their application in the robotics field, along with fundamental models~\cite{driess2023palm}\cite{codeaspolicies2022}.
In particular, it is possible to generate action planning for the robot task using complex natural language instructions as input, which has been difficult in past studies~\cite{Ahn2022}\cite{Brohan2022}.
LLMs can control their outputs by arbitrarily designing the prompts they are given as input~\cite{OpenAI}\cite{kojima2022large}.
In robot action planning, the prompts are usually designed to output primitive action commands from the collaborator's instructions.
However, because of ambiguity or lack of context-dependent information in the natural language, the LLM's response may not be the one that the collaborator desires.
It is possible to adjust the output of the LLM by elaborating the action instruction in more detail, but the design cost is high and impractical.
In this study, we aim to realize a clear action plan by utilizing interactive prompt engineering to address the above issues.

\begin{figure}[tb]
    \centering
    \includegraphics[width=\columnwidth]{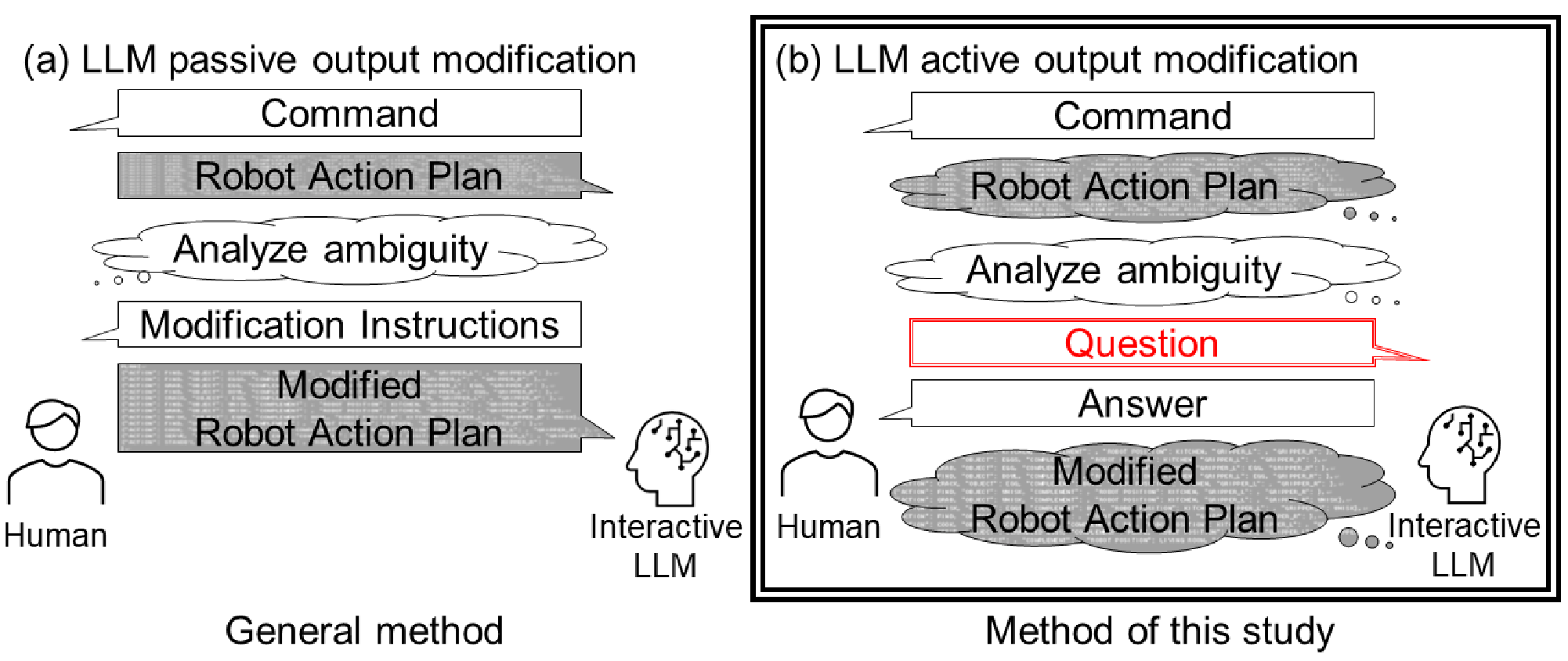}
    \caption{
Two interactive output modification  methods.
(a) LLM passive output modification  by elaborate prompts.
(b) LLM active output modification  by questioning collaborators.
    }
    \label{fig:intro}
\end{figure}

When modifying LLM output in robot action planning, it is common to be based on unidirectional instructions from the collaborator (Fig.~\ref{fig:intro}a).
The collaborators point out the incorrect part of the action command presented by the LLM, and re-enter the prompt to correct it.
In this case, the LLM only receives instructions passively.
Although the above method can accurately reflect the collaborator's intentions, it is burdensome because correction instructions must be issued each time.

On the other hand, from the perspective of active inference~\cite{Suzuki2023} to reduce the prediction error of the model, it is important to use a method in which the LLM actively modifies its output (Fig.~\ref{fig:intro}b).
Through dialog with the external world, we work to reduce the gap between the LLM and the real world.
This paper makes the LLM evaluate the uncertainty of its own planning output of robot tasks, and ask questions to collaborators regarding the modification  points.
Since the LLM automatically provides the corrective actions, the ambiguity of the action instructions is expected to be resolved by the collaborators' answers to the questions, thereby reducing the cost of designing the initial action instructions.

Based on the above background, we focus on (b) active output modification  of LLM, and study to clarify robot action plans.
In the proposed method, the LLM interactively modifies its output by presenting information that is missing in relation to the final planning result.
We use ChatGPT~\cite{ChatGPT} as an interactive LLM, and verify the effectiveness of the above method through interactive experiments using examples of the robot cooking task.

\section{Related Works}

The LLMs have been used in many robotics studies, and the use of multi-modalities with foundational models has also attracted attention.
Kawaharazuka et al. used a vision-language model for robot environment recognition and proposed a method to improve recognition accuracy through visual question answering tasks~\cite{Kawaharazuka2023}.
Ma et al. showed that better reward expressions can be obtained by applying the vision-language model to reinforcement learning~\cite{Ma2023}.
These studies suggest that the expressive ability of LLMs is applicable to real robot tasks.

On the other hand, in robot motion planning, it is possible to convert task instructions by humans into action sequences in JSON format~\cite{Ahn2022}\cite{huang2022language} or programming code format~\cite{codeaspolicies2022}\cite{Vemprala2023}.
Among them, Brohan et al. have attempted to ground robot actions with language instructions and realized simple robot tasks in real environments~\cite{Brohan2022} using a model that has learned an action planning scheme in advance~\cite{Ahn2022}.
While these studies relate to the clarification of robot action planning, they do not consider the process of interactive modification .
Vemprala et al. reported an example of using an interactive LLM to generate a robot action plan at the programming code level and adjust it while checking and providing feedback using a simulator~\cite{Vemprala2023}.
However, since there are no modification  suggestions from the LLM side, the collaborators need to design the details of the modification s to the LLM.

The contributions of this study are summarized as follows:
\begin{itemize}
    \item We propose an interactive robot action planning method based on uncertainty analysis and active questioning by LLM.
    \item We verify the effectiveness of the proposed method through dialogue experiments of multiple cooking tasks.
\end{itemize}


\section{Method}
\begin{figure}[tb]
    \centering
        \includegraphics[width=\columnwidth]{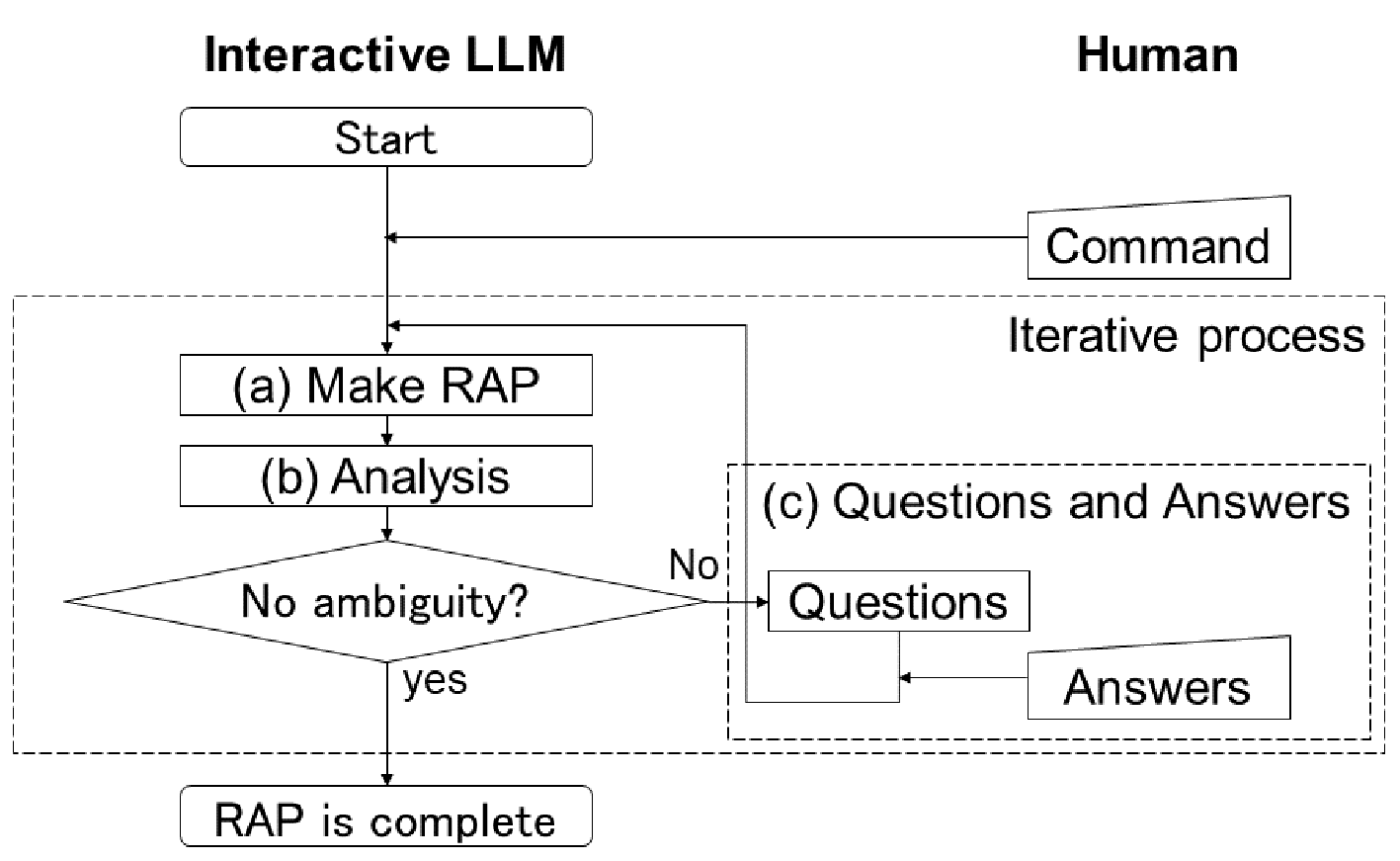}
    \caption{
    Overview of the proposed method
    }
    \label{fig:method}
\end{figure}

In this study, we design prompts to be given to the LLM for robot action planning.
Fig.~\ref{fig:method} shows an overview of the process of the proposed method.
After the Command information indicating robot task instructions is given from the human to the LLM, the output is interactively modified by iterating the following three processes.

\begin{description}
\item[(a) Make Robot Action Plan]\mbox{}\\
The LLM compiles the robot's action plan into a unique format and outputs it.
In this paper, this sequence of actions is called Robot Action Plan (RAP).
This process is realized by the following description in the prompt.
"a) Make RAP (provide a modified  RAP. It should be something that the robot can easily understand. Therefore, the prompt should be unambiguous.)
a-1) RAP should be output as a list. 
RAP is generated based on the Command first given to LLM and the information obtained from Questions and Answers.
However, for the first time, RAP is generated using only the information in the Command.

\item[(b) Uncertainty Analysis]\mbox{}\\
LLM analyzes the ambiguity in the generated RAP and outputs the analysis results in text.
Here, ambiguity is considered as uncertainty in the prediction of LLM, and its analysis is based on the LLM's subjectivity.
This process is realized by the following description in the prompt.
"Please analyze step by step what elements are missing in the RAP for the robot to work. Then output the information that should be added to the RAP. If there is no information to be added, please output 'none'."
Once the LLM determines that the RAP is clear in this analysis, the entire process is terminated.

\item[(c) Questions and Answers]\mbox{}\\
If the ambiguity is still in the generated RAP, the LLM and humans interact to collect additional information.
The LLM generates questions to disambiguate the RAP, and humans then enter their answers.
The questions are based on the results of analysis in (b) Analysis.
This process is realized by the following description in the prompt.
"Please collect the information you suggested in the b) analysis that should be added to the RAP by asking questions. I will provide the information for your question. If you have no questions, please output 'none'."
Note that multiple questions may be output at the same time, and the human can input multiple answers also at the same time.
The human can refuse to answer questions that are difficult to answer.
The information obtained in this process is used to generate RAP again, and the proposed processes (a-c) are repeated until the final RAP is output.

\end{description}

\begin{figure*}[t]
    \centering
    \includegraphics[width=2.0\columnwidth]{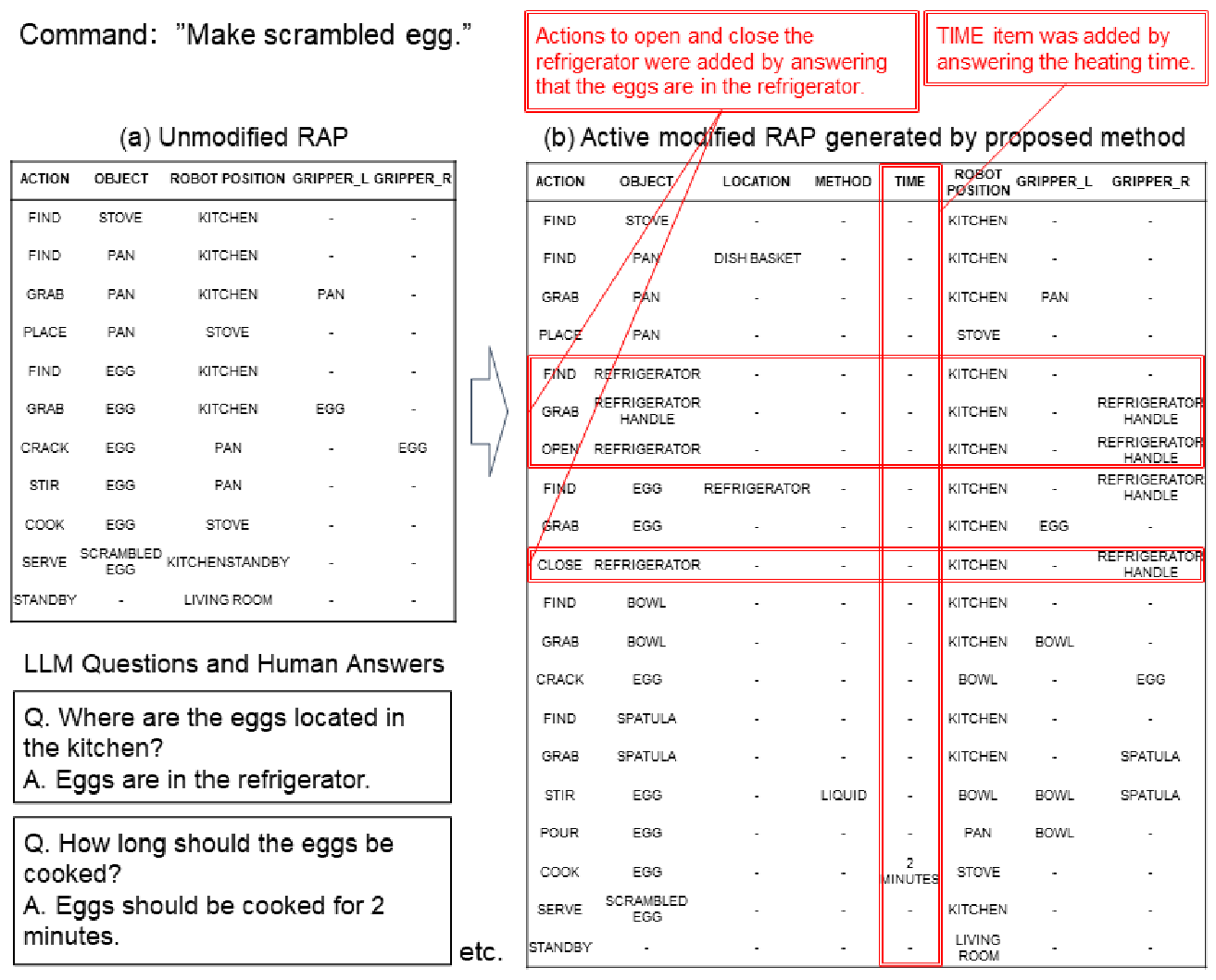}
    \caption{Comparison results of RAPs before and after applying the proposed method to Task 1 (Make scrambled egg) in Experiment 1.}
    \label{fig:result1-1}
\end{figure*}

\begin{figure*}[t]
    \centering
    \includegraphics[width=2.0\columnwidth]{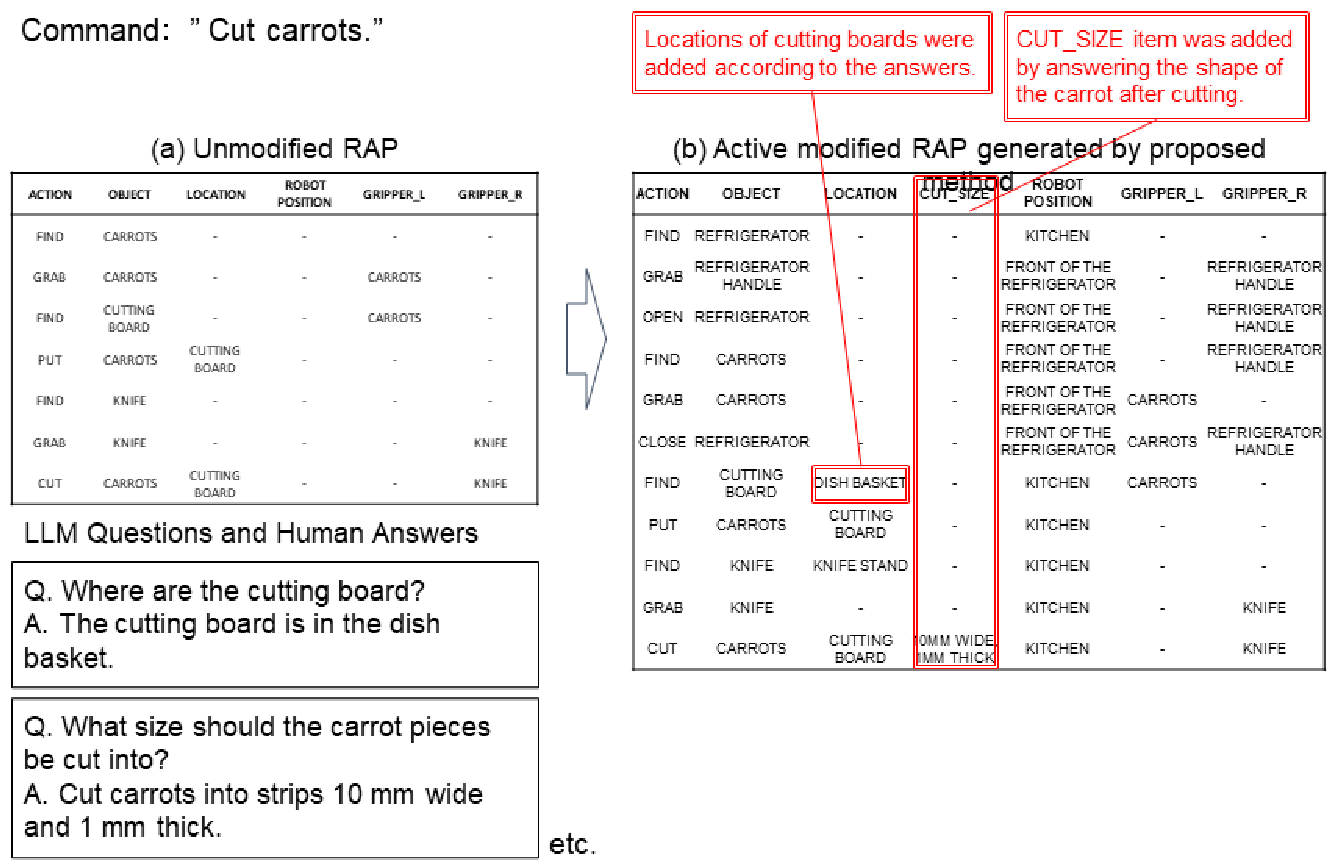}
    \caption{Comparison results of RAPs before and after applying the proposed method to Task 2 (Cut carrot) in Experiment 1.}
    \label{fig:result1-2}
\end{figure*}

\section{Experiments}
To verify the effectiveness of the proposed method, a dialogue experiment is conducted for outputting RAPs from commands given by humans.
We will investigate the changes in RAP and the content of the questions asked through the dialogue for the action plans of multiple cooking tasks.

\subsection{Prompt Engineering}
We used Chat-GPT (version gpt-4-0314) released by OpenAI, and its parameter of temperature was set to 0.
The prompts provided to the LLM consist of the following parts: "Role" for the role of the LLM and purpose, "Prerequisites" for the preconditions of the task and robot, "Process" for the processing procedure, "Output" for the RAP format, and "Example" for concrete examples of LLM's input and output.
As the Prerequisites part, only the conditions related to the robot's ability and the initial state of the task (e.g., position) are described, as shown below.
\begin{enumerate}
   \item The robot has two robotic arms.
   \item The robot arm has 7 degrees of freedom.
   \item The robot can grab things at will.
   \item The robot can acquire information about the appearance of objects by means of a camera.
   \item The robot has a pre-mapped information of the workspace.
   \item The robot is currently in the living room.
   \item The human (MASTER) who gives commands to the robot is sitting on a chair in the living room.
\end{enumerate}
The Process part is constructed with the prompts shown in the previous section.

The output format used in this experiment is RAP, which is designed in JSON format to represent the robot's action plan in a comprehensive and structured manner.
Our RAP consists of "ACTION" indicating the motion class, "OBJECT" indicating the target of the motion, "ROBOT POSITION" indicating the robot's location, and "GRIPPER\_L" and "GRIPPER\_R" indicating what the arm is holding.
In addition, we recommend adding RAP items to the LLM at its own discretion by adding the statement; "It is recommended to add formatting items as needed."
As concrete examples, we input RAPs for the tasks of "getting an energy drink from the refrigerator in the kitchen" and "making banana milk. 
The total amount of tokens in the prompts was about 4150.

\subsection{Evaluation}
As evaluation tasks, we selected a cooking task by robots.
Cooking tasks are required for general-purpose household robots and have been widely studied because many of them have complex work processes~\cite{Dong2021}\cite{Liu2022}.
Since tasks with complex work processes are generally required a high design cost of motion instructions, the cooking tasks are suitable to confirm the effectiveness of the proposed method. This experiment involved cooking scrambled eggs (Task 1) and cutting carrots (Task 2). The commands given to LLM are "Make scrambled egg." and "Cut carrots," respectively. The above was designed simply to include a large amount of ambiguity regarding the cooking process and cooking environment.
We will test whether this ambiguity can be clarified using LLM's active output modification .
Three trials were conducted for each task.

We verify the effectiveness of the proposed method by conducting comparative experiments on the above robot tasks from two perspectives.
First, we compare the RAPs before and after applying the proposed method (Experiment 1).
By comparing the RAP generated in the initial process (a), which does not go through an iterative process, and the final RAP, we confirm whether the RAP is improved by uncertainty analysis and active questioning.

We also verify the difference in the quality of the generated RAPs by varying the amount of information given to the LLM in the initial phase of motion planning (Experiment 2).
By comparing the RAP generated based on a human-elaborated Command and the RAP generated by the proposed method based on a simple Command that contains ambiguity, we will examine whether the modification s proposed by the LLM are appropriate for the robot task.

\section{Results and Discussion}
\subsection{The RAPs before and after applying the proposed method}
The results of Experiment 1 are shown in Fig.~\ref{fig:result1-1} and Fig.~\ref{fig:result1-2}. 
Tables (a) and (b) in each figure summarize the RAPs before and after applying the proposed method. 
The final RAP was determined by the LLM to be sufficiently clear, following the interaction examples (question and answer) about the location of the eggs and the cooking time of the eggs shown in the lower left of the figure. 
In this experiment, the average number of questions output by the LLM to complete the RAP was 2.66 for both Tasks 1 and 2. 
Some of the questions were designed to elicit multiple pieces of information at once. 
The average number of iterative processes required to output RAPs for Tasks 1 and 2 were 2.33 and 2.00, respectively.

Comparing the tables in Task 1 (Fig.~\ref{fig:result1-1}), it can be seen that the information obtained through the interaction is reflected in the RAP. 
It was confirmed that the actions to open and close the refrigerator were added by responding that the eggs were in the refrigerator, and that a TIME item was added by responding the heating time. 
In particular, the latter can be said to show performance beyond the predefined output format, indicating that the proposed method is effective in improving the content of the RAP.

In Task 2 (Fig.~\ref{fig:result1-2}), it can be seen that the information obtained through the interaction is reflected in the final RAP, similar to the results of Task 1. 
The LOCATION information was added to the action to find the cutting board by answering that the cutting board is in the basket, and the CUT\_SIZE item was added by answering the size of the carrot to be cut.

\begin{figure*}[t]
    \centering
    \includegraphics[width=2.0\columnwidth]{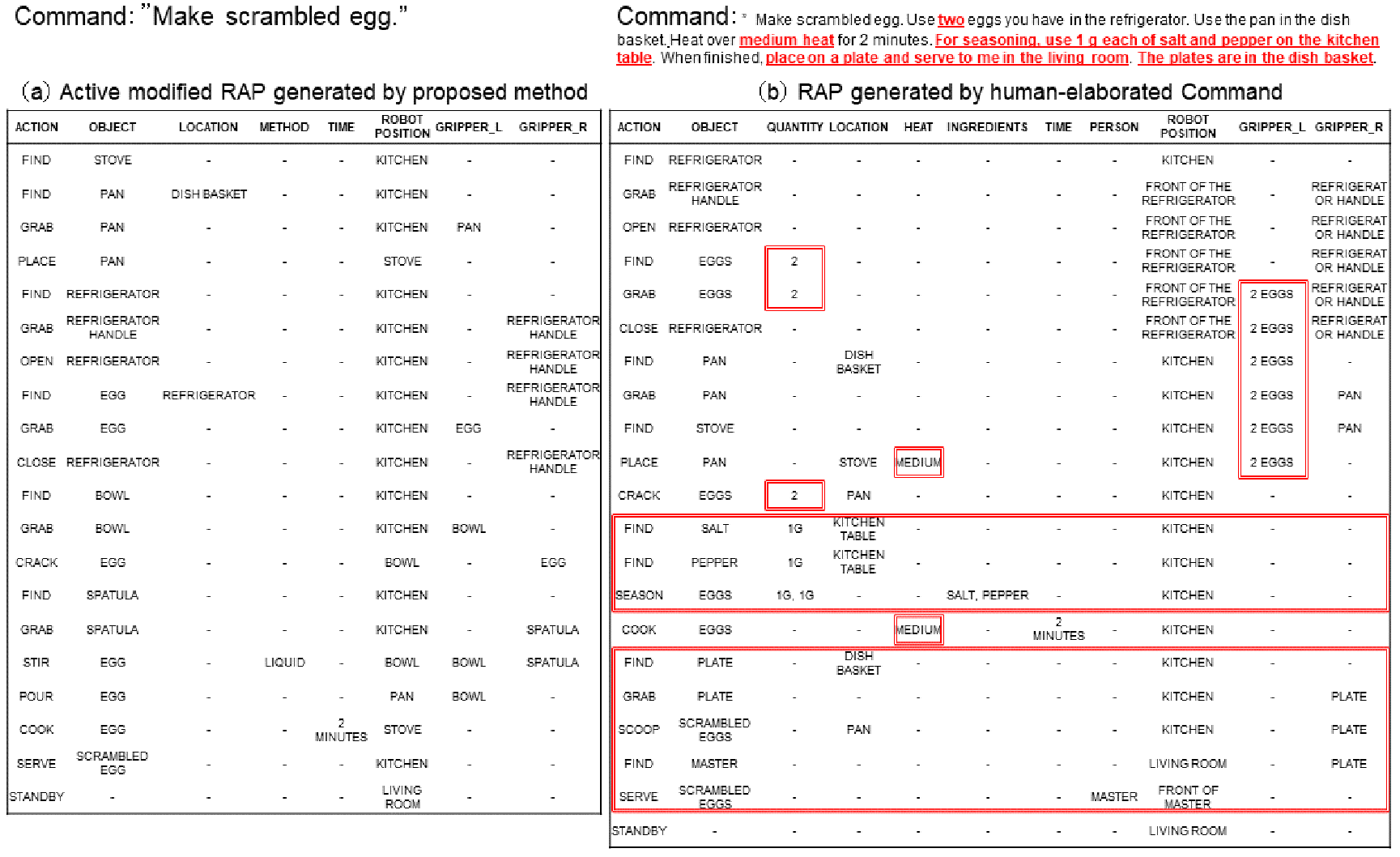}
    \caption{
Comparison of RAPs for Task 1 (Make scrambled egg) in Experiment 2, based on changes in the amount of Command information. The red frame indicates information that is not included in the RAP of (a) but is included only in the RAP and Command of (b).
}
    \label{fig:result2-1}
\end{figure*}

\begin{figure*}[t]
    \centering
    \includegraphics[width=2.0\columnwidth]{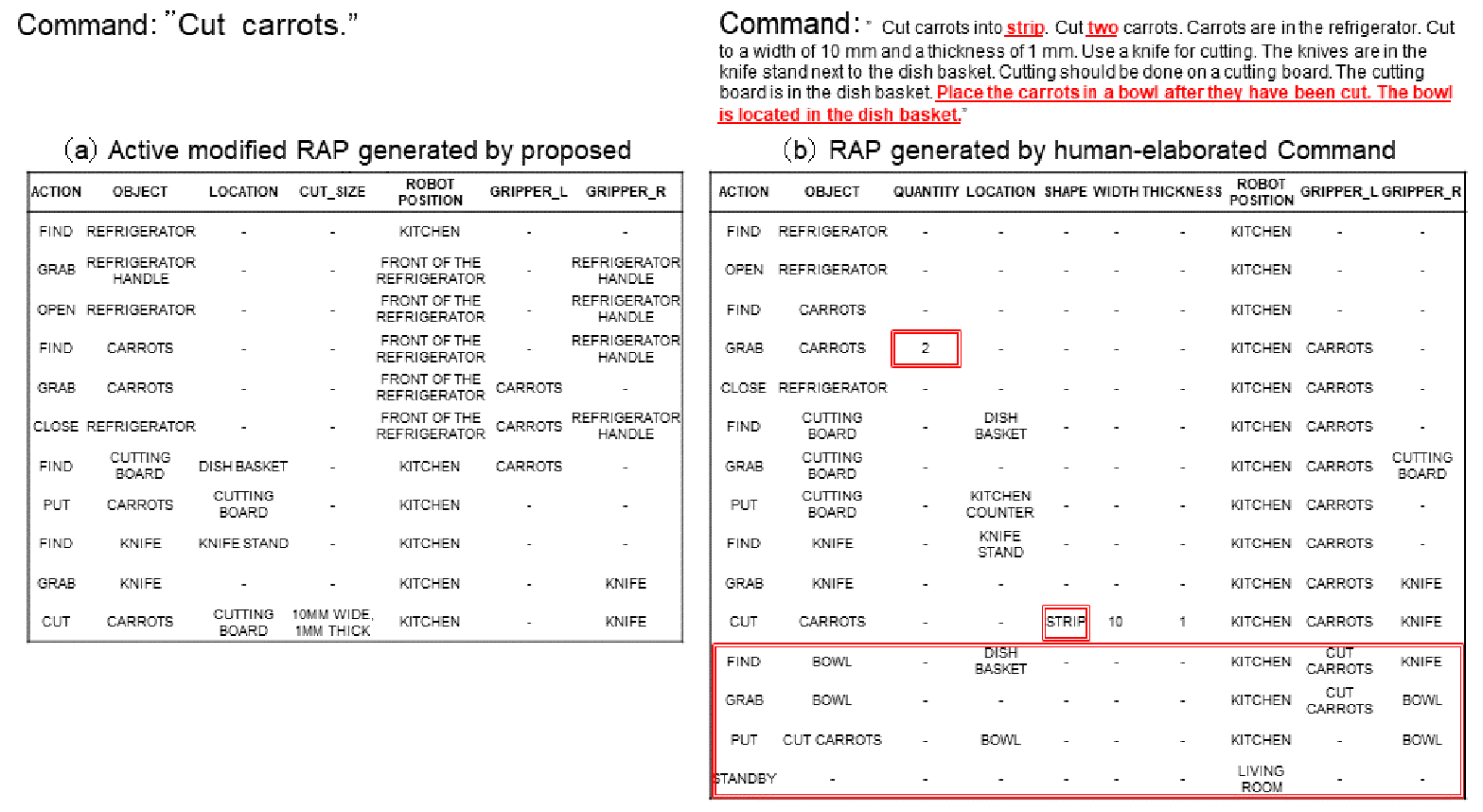}
    \caption{Comparison of RAPs for Task 2 (Cut carrots) in Experiment 2, based on changes in the amount of Command information. The red frame indicates information that is not included in the RAP of (a) but is included only in the RAP and Command of (b).
}
    \label{fig:result2-2}
\end{figure*}

\subsection{Comparison by changing the Command information}
The results of Experiment 2 are shown in Fig.~\ref{fig:result2-1} and Fig.~\ref{fig:result2-2}.
Table (a) in each figure shows the final RAP obtained by the proposed method in Experiment 1, and Table (b) shows the RAP generated by the human-designed command.

Comparing the tables in Task 1 (Fig.~\ref{fig:result2-1}), it can be seen that the RAP in Table (a) does not contain some information that the RAP in Table (b) does.
They are information that affects the quality of the scrambled eggs, such as the number of eggs, the heat level, and the seasoning, as well as the serving information. 
Both of the above are useful, but not essential, in making scrambled eggs. 
It is suggested that the proposed method can generate as much RAP as a detailed manual Command by prioritizing information that is important for action planning. 
In addition, the missing information described above was input to the elaborated Command (passive output modification ), and we can expect to obtain a better RAP by using both passive and active output modification .

Finally, comparing the tables in Task 2 (Fig.~\ref{fig:result2-2}), it can be seen that the RAP in Table (a) does not contain some information that the RAP in Table (b) does. 
They are the shape of the carrots after cutting, the number of carrots, and information about the arrangement of the carrots. 
This information is also useful, but not essential, in cutting carrots. 
Combined with the results of Experiment 1, the effectiveness of the proposed method for accurate RAP generation was demonstrated.

\subsection{Limitation and Discussion}
While the effectiveness of the proposed method was demonstrated through Experiments 1--2, several issues remain.
The issues described below may be controlled by the design of the prompts, but they are not limited to robot motion planning. The first is the amount of tokens to be input to the LLM.
The proposed method requires a large number of sentences explaining the process in the prompt, which increases the overall token volume.
The large amount of tokens slows down the response speed of the LLM and squeezes the space to describe specific examples.
This is an important technical issue to be solved in motion planning using interactive LLMs, and is often caused by the computational cost of the attention mechanism of the Transformer-type model.
~\cite{choromanski2020masked}\cite{katharopoulos2020transformers}.
In addition, the realization of complex processes only through prompts alone is limited, so an adaptive process of output changes in the real world is required.

Second, there is the issue of asking unimportant questions.
For example in our experiment, the question about the type of egg was asked in Task 1.
The above question may be important in certain situations, but it is not that important in the general task of making scrambled eggs.
Thus, since the importance of the question changes depending on the task context, it is necessary to examine the model including the robot control model in the latter stage.

Finally, there are cases in which the interpretation in the LLM takes a leap.
For example, when a robot is given the task of cutting carrots, it sometimes outputs a RAP for making a salad.
Since the above is done without human confirmation, it may cause unexpected problems when considering implementation in a robot.
This is also related to the hallucination problem~\cite{guerreiro2023hallucinations}.
It could be manipulated by intervening in the LLM reasoning process but is essentially a difficult problem to solve.

The latter two issues are largely due to the fact that the robot's body is not taken into account when learning LLMs, and it is important to connect LLM recognition to the real environment with limited data~\cite{Toyoda2021}\cite{Toyoda2022}.
We believe that integration with robot hardware will be necessary, and we plan to work on these issues in the future.


\section{Conclusion}
In this study, we investigated a method for modifying the robot’s action plan based on the LLM’s self-analysis and question process. Dialog experiments on two types of cooking tasks, cooking scrambled eggs and cutting carrots into strips, show that the proposed method can reduce the design cost of the instructions given to the LLM. However, it was also shown that the proposed method has several challenges caused by the LLM’s lack of information about the real environment, such as unimportant questions and leaps in action planning. In the future, we will work on the development of a more realistic action planning method from the viewpoint of integrating LLM and robot motion generation models.

\section*{ACKNOWLEDGMENT}
This work was supported by JST Moonshot R\&D Grant Number JPMJMS2031 and Research Institute of Science and Engineering, Waseda University.

\bibliographystyle{IEEEtran}
\bibliography{root}

\end{document}